\documentclass[conference]{IEEEtran}
\IEEEoverridecommandlockouts

\title{Learning Displacement-Robust Representations for Landslide Early Warning under Rainfall Forecast Uncertainty}

\def\BibTeX{{\rm B\kern-.05em{\sc i\kern-.025em b}\kern-.08em
    T\kern-.1667em\lower.7ex\hbox{E}\kern-.125emX}}

\usepackage{graphicx}
\usepackage{textcomp}
\usepackage{xcolor}
\usepackage{bm}
\usepackage{soul}
\usepackage{flushend}
\usepackage{xspace}
\usepackage{soul}
\usepackage{algorithm}
\usepackage{algpseudocode}
\usepackage {diagbox}

\usepackage{multirow} 
\usepackage{booktabs}
\usepackage{tcolorbox}
\usepackage{hhline}

\usepackage{booktabs}
\usepackage{multirow}
\usepackage{graphicx}
\usepackage{threeparttable} 
\usepackage{xcolor}
\usepackage{colortbl}
\usepackage{amsmath} 
\usepackage{amssymb}

\usepackage{url}

\author{
\IEEEauthorblockN{
Ren Ozeki\textsuperscript{†\P}, 
Hamada Rizk\textsuperscript{†‡§}, 
Hirozumi Yamaguchi\textsuperscript{†§}
}
\IEEEauthorblockA{\textsuperscript{†}Osaka University, Suita, Japan}
\IEEEauthorblockA{\textsuperscript{‡}RIKEN Center for Computational Science, Kobe, Japan}
\IEEEauthorblockA{\textsuperscript{§}Tanta University, Tanta, Egypt}
\IEEEauthorblockA{\textsuperscript{\P}Japan Meteorological Agency}
\IEEEauthorblockA{
\texttt{\{r-ozeki, hamada\_rizk, h-yamagu\}@ist.osaka-u.ac.jp}
}
}

\begin{document}
\maketitle

\begin{abstract}
Rainfall-induced landslides pose a growing risk worldwide as climate change intensifies extreme rainfall events.
To provide sufficient time for evacuation, landslide early warning systems (LEWS) used for real-time disaster monitoring must estimate near-future landslide risk by integrating observed rainfall with short-term rainfall forecasts from spatio-temporal environmental data streams.
Although recent landslide prediction methods have improved predictive performance using statistical and deep learning approaches, most assume accurate rainfall inputs.
In operational settings, however, landslide prediction relies on rainfall forecasts, which often contain spatial displacement of rainfall fields due to forecasting uncertainties.
Such displacement can alter local accumulated rainfall and degrade landslide prediction accuracy.
To address this challenge, we propose a novel LEWS that is robust to rainfall field displacement.
The key idea is to learn latent representations from rainfall and terrain data that remain stable under displacement in rainfall field motion, enabling reliable geospatial data integration for landslide risk estimation.
The landslide prediction model is trained using Rainfall-Motion-Aware Contrastive Learning (RMCL), which introduces temporally correlated rainfall field perturbations to emulate forecast-induced displacement in rainfall-driven spatio-temporal environmental data streams.
Experiments were conducted using two years of rainfall and terrain data across Japan, covering 19 regions where landslides occurred.
The proposed system achieved up to 37\% higher precision than state-of-the-art baselines.
These results demonstrate that modeling rainfall as a moving spatial field and addressing rainfall field displacement within the learning process significantly improve the reliability of short-term landslide prediction in operational early warning systems for real-time disaster monitoring.
\end{abstract}

\maketitle

\section{Introduction}

\begin{figure*}
    \centering
    \includegraphics[width=1.0\linewidth]{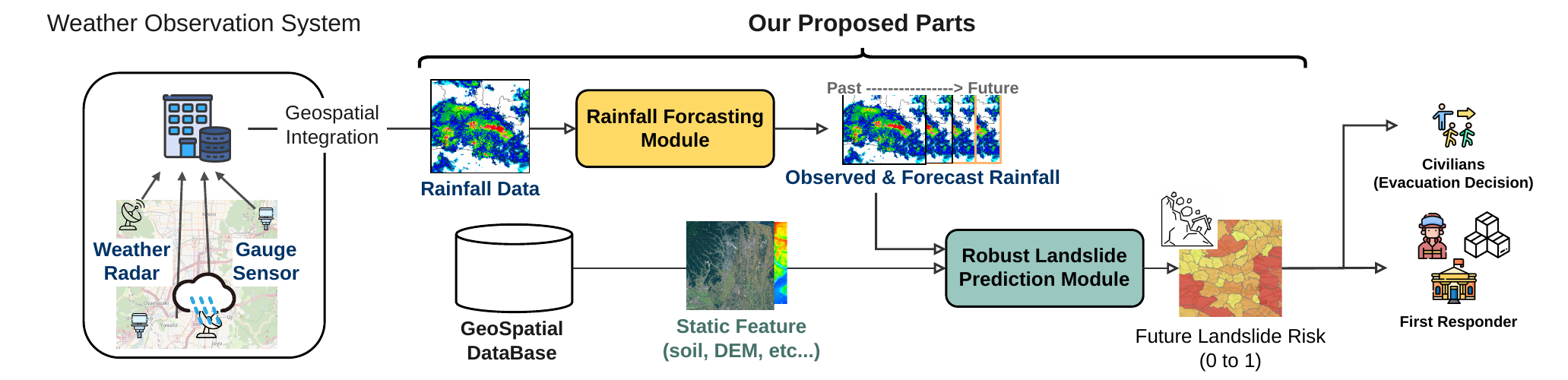}
    \caption{The overview of the landslides early warning system.}
    \label{fig:sys_overview}
\end{figure*}

Global climate change has intensified extreme rainfall events and increased the frequency of rainfall-induced landslides worldwide \cite{newman2023global,world2022state}. 
Rainfall-induced landslides occur when rainfall accumulates in the soil, increasing soil moisture and reducing slope stability \cite{kuramoto2001study}. 
Therefore, monitoring rainfall as part of spatio-temporal environmental data streams is critical for predicting landslide occurrence \cite{kuramoto2001study, mondini2023deep}. 
To provide sufficient time for residents to evacuate safely, early warning systems that assess landslide risk before failure occurs are an essential component of real-time disaster monitoring within disaster management systems \cite{casagli2023landslide}. 
Thus, in operational settings, landslide early warning systems (LEWS) commonly estimate future landslide risk by considering not only current and preceding rainfall conditions but also the expected rainfall over the next several hours \cite{casagli2023landslide}, as shown in Figure~\ref{fig:sys_overview}.

In such LEWS, rainfall observations collected from weather radars and rain gauges form large-scale spatio-temporal environmental data streams that support geospatial data integration and are continuously updated as time series on a spatial grid.
These data represent the spatial distribution of rainfall as a rainfall field.
Over short timescales of several hours, rainfall fields evolve through rainfall field motion, where coherent rainfall regions move across the landscape driven by atmospheric flow \cite{ritvanen2023advection}. 
To estimate future landslide risk, rainfall forecasting models generate short-term rainfall forecasts that are incorporated into LEWS in the same spatio-temporal format as observed rainfall data for real-time disaster monitoring. 
However, forecast errors are not avoidable and often appear as spatial displacements of the rainfall field. 
Such displacement changes whether and how long rainfall remains over a particular location. 
Since rainfall forecasts are directly used to compute accumulated rainfall and other indicators in landslide prediction models, displacement of the rainfall field can alter the inputs to landslide prediction models and consequently affect prediction accuracy.
Additionally, since landslides are extreme events, the number of occurrence cases is limited, resulting in a scarce and highly imbalanced dataset. 
Such characteristics of landslide data hinder landslide prediction models from acquiring comprehensive knowledge of landslide mechanisms and reduce their robustness to input perturbations caused by rainfall forecasting. 
Due to these two challenges, accurate future landslide prediction under operational settings remains difficult.

Previous studies on landslide prediction have explored statistical and machine learning models based on rainfall history and terrain conditions \cite{geosciences9070302, ho2017performance, nguyen2019development, liu2020algorithms}. 
Representative approaches include rainfall threshold models that estimate triggering conditions using the relationship between rainfall intensity and duration, as well as machine learning-based risk estimation methods that extend these frameworks \cite{casagli2023landslide, kuramoto2001study}. 
More recently, predictive performance has been improved by integrating high-resolution rainfall data and various environmental variables with CNNs or Transformers \cite{ge2024litetransnet, zhao2024landslide}. 
However, many studies assume that the rainfall inputs used in landslide prediction models are accurate. 
Consequently, although these models perform well when evaluated using observed rainfall data, their predictive performance may degrade when forecast rainfall is used as input, as investigated in \cite{khan2021investigating}. 
To address spatial displacement in rainfall forecasts, some studies train landslide prediction models directly using forecast rainfall data \cite{wang2024spatio, chen2024attention}. 
However, since these models rely on forecast rainfall, they may overfit to forecast-specific features rather than the underlying mechanisms of rainfall-induced landslides. 
Furthermore, accumulated forecast errors over longer prediction horizons can further degrade model robustness and predictive performance.

To address these challenges, we propose a novel landslide early warning system designed to maintain predictive performance under rainfall forecast uncertainty. 
The key idea is to learn latent representations from rainfall and terrain data that are robust to displacement in rainfall field motion.
To achieve this, we model rainfall as a moving spatial field over grid-based spatio-temporal data and emulate plausible displacement of rainfall fields using a temporally correlated perturbation process. 
These augmented rainfall fields are used in Rainfall-Motion-Aware Contrastive Learning (RMCL) to learn representations that remain stable under the displacement of rainfall field motion. 
By integrating rainfall field perturbation with representation learning, the proposed approach enables landslide prediction models to maintain stable performance even when rainfall forecasts contain spatial displacement errors. 
Furthermore, the proposed contrastive learning enables landslide prediction models to learn meaningful representations by utilizing pairwise relationships among samples within the same class, allowing them to acquire comprehensive knowledge even from imbalanced data.

To evaluate the proposed framework, we collected two years of rainfall and terrain data across Japan, including 19 regions where landslides occurred. 
Using a chronological train-test split to prevent time-series leakage and reflect operational forecasting conditions, 70\% of the earlier data were used for training and the remaining 30\% for testing. 
Model performance was assessed using precision at 80\% recall to account for severe class imbalance. 
Experimental results demonstrate that the proposed system achieves up to 43\% higher precision compared with state-of-the-art baselines. 
These findings highlight the importance of modeling rainfall as a moving spatial field and addressing rainfall field displacement within the learning process for reliable short-term landslide prediction.

\section{Related Work}
\label{sec:related_work}
\subsection{Landslide Prediction Systems}

Recent global climate change has led to more frequent disasters, even in previously low-risk areas, increasing the demand for accurate and reliable landslide prediction.
Traditional landslide prediction approaches are algorithm-based methods \cite{geosciences9070302, ho2017performance, nguyen2019development,liu2020algorithms}.
For instance, the authors of \cite{geosciences9070302} proposed a statistical method to predict landslides and evaluated their method using a dataset from India as a case study.
However, these approaches do not sufficiently consider the spatio-temporal dependency of landslides, such as soil distribution, elevation, and rainfall-driven water accumulation.
As a result, these methods have significant limitations in predictive accuracy.

To capture the spatio-temporal dependency of landslides, machine learning methods have also been proposed \cite{collini2022predicting, xie2019application, ge2024litetransnet}.
The authors of \cite{collini2022predicting} predicted landslide events using CNNs to capture spatial dependencies.
The work in \cite{xie2019application} predicted ground displacement, which can be an indicator of landslides, by utilizing LSTM models.
Although these approaches capture the spatio-temporal dependency of landslides, they typically assume that accurate rainfall observations are available.
In operational LEWS, however, future rainfall must be provided by nowcasting or short-term forecasts, which contain uncertainties in position and intensity \cite{ritvanen2023advection}.
As a result, models developed under such idealized conditions may suffer performance degradation when applied with forecast rainfall that includes spatial displacement errors \cite{chen2024attention, khan2021investigating}.

To overcom spatial displacement in rainfall forecasts, several studies have trained landslide prediction models using forecast rainfall data \cite{wang2024spatio, chen2024attention}.
While this approach enables models to incorporate forecast uncertainty during training, the reliance on forecast rainfall may cause the models to learn forecast-specific characteristics rather than the underlying mechanisms of rainfall-induced landslides.

\textit{
In contrast, the proposed system addresses this gap by incorporating Rainfall-Motion-Aware Contrastive Learning (RMCL).
Specifically, we adopt supervised contrastive learning with rainfall field perturbation to emulate forecast-induced displacement of rainfall fields.
By training the model to produce consistent representations under such controlled perturbations, the proposed approach improves robustness to rainfall field displacement and enhances landslide prediction performance in operational settings.
}

\subsection{Class Imbalance in Landslide Prediction}
Landslide early warning systems face severe class imbalance because landslide events are rare compared to non-event conditions. 
Since standard classifiers typically optimize loss functions that treat classes equally, they tend to be biased toward the majority class, reducing sensitivity to hazardous events. 
However, in landslide prediction, correctly identifying these rare minority events is of primary importance.
Existing approaches to class imbalance can be broadly categorized into three groups: data-level, algorithm-level, and ensemble methods \cite{alsaui2022resampling, singh2022credit}.
Data-level approaches modify the training dataset to balance class ratios before model training. 
Typical strategies include undersampling \cite{9533379} and oversampling \cite{chawla2002smote, dablain2022deepsmote}. 
While undersampling may remove informative majority samples, oversampling techniques such as SMOTE\cite{chawla2002smote} generate synthetic minority samples and have been widely applied in traditional machine learning models. 
However, SMOTE-based methods struggle with multi-modal data distributions and high-dimensional feature spaces. 

Algorithm-level approaches directly modify the learning process of the classifier. 
Common strategies include loss re-weighting, cost-sensitive learning \cite{9064578}, Mean False Error \cite{7727770}, and Focal Loss \cite{lin2017focal}. 
Focal Loss reduces the contribution of easy samples and emphasizes hard or minority examples, making it particularly suitable for rare-event detection. 
Ensemble approaches combine multiple classifiers trained on rebalanced datasets \cite{LIU201735}, and some methods integrate ensemble learning with undersampling \cite{DIEZPASTOR201596} to improve minority detection performance.
Despite these advances, a severe imbalance combined with limited positive samples remains challenging for deep neural networks. 
When minority samples are extremely scarce, learning stable and meaningful representations becomes difficult, and models may overfit to regional or dataset-specific patterns.

\textit{
To address this issue in LEWS, our system integrates supervised contrastive learning as pretraining method for the landslide prediction model. 
This approach enables landslide prediction models to learn meaningful representations using pairwise relationships among samples within the same class, allowing them to acquire comprehensive knowledge even from imbalanced data.
}

\section{Proposed System}
Figure~\ref{fig:sys_overview} illustrates the overall landslide disaster management system and highlights the components addressed in this study.
Rainfall observations collected from multi-sensor radar of weather observation systems are first integrated into a unified geospatial space to produce rainfall field data.
The proposed system, corresponding to LEWS, operates on this rainfall data to enable short-term landslide prediction in operational settings where future rainfall is not directly observable.
The proposed LEWS consists of two components: a rainfall forecasting model and a landslide prediction model.
The rainfall forecasting model performs short-term rainfall forecasting to generate a continuous sequence of observed and forecast rainfall fields.
These rainfall fields are subsequently used by the landslide prediction model to estimate landslide risk and support evacuation decisions and first-responder actions.

First, the rainfall forecasting model generates short-term rainfall forecasts by modeling rainfall field motion.
Using an optical-flow-based approach, the model estimates the horizontal motion of the rainfall field and extrapolates the most recent observation to future lead times.
This design reflects the dominant transport mechanism of short-term rainfall\cite{ritvanen2023advection} and allows the system to operate in practical early warning settings where actual future rainfall is unavailable.
The predicted rainfall fields serve as direct inputs to the subsequent landslide prediction model.

Second, the landslide prediction model integrates heterogeneous information sources to estimate landslide probability.
It combines (i) observed and forecast rainfall fields and (ii) geographically anchored spatial features such as terrain and soil properties.
Since landslide prediction is sensitive to the displacement of the rainfall field relative to terrain, the model is pretrained using Rainfall-Motion-Aware Contrastive Learning (RMCL).
RMCL introduces rainfall field perturbation to emulate forecast-induced displacement of rainfall fields and trains the model to learn representations that remain stable under such displacement.
The following subsections describe each component in detail.

\subsection{Rainfall Forecasting Model}
\label{subsec:rainfall_module}
The rainfall forecasting model generates short-term rainfall forecasts that serve as inputs to the landslide prediction model. 
Since short-term rainfall evolution is primarily governed by rainfall field motion\cite{ritvanen2023advection}, we adopt an optical-flow-based approach to estimate the horizontal motion of the rainfall field and extrapolate the most recent observation forward in time.
Specifically, we compute the motion field by applying a dense optical flow method to a short time series of rainfall fields. 
We employ the dense Lucas–Kanade algorithm \cite{bouguet2001pyramidal}, which is widely used for radar-based rainfall nowcasting \cite{pulkkinen2019pysteps, ritvanen2023advection}. 
The motion of the rainfall field is estimated from the three most recent rainfall fields, allowing the model to capture short-term rainfall field motion.
Using the estimated motion field, the latest observed rainfall field is advected forward to generate rainfall forecasts up to eight hours ahead. 
At this forecast horizon, spatial displacement of the rainfall field typically dominates rainfall evolution, making advection-based extrapolation appropriate. 
The predicted rainfall fields are then supplied to the landslide prediction model, enabling deployment in operational settings where future rainfall observations are unavailable.

\subsection{Landslide Prediction Model}

\subsubsection{Model Architecture}
The proposed landslide prediction model takes two types of inputs: 
(1) geographically anchored spatial features such as terrain and soil properties, and 
(2) spatio-temporal rainfall fields.  

Spatial features are extracted using a two-layer CNN that captures local terrain patterns.  
Rainfall sequences are processed by a CNN–Transformer composed of a two-layer CNN and a three-layer Transformer to model temporal dependencies and complex rainfall interactions.  
The resulting spatio-temporal features are summarized via mean pooling to form a fixed-length representation.  
Finally, the spatial features and pooled spatio-temporal rainfall features are concatenated into a unified feature vector used by downstream prediction heads (e.g., binary landslide classification or contrastive learning).

\subsubsection{Rainfall-Motion-Aware Contrastive Learning (RMCL)}
\label{subsec:sacl}
\begin{figure}
    \centering
    \includegraphics[width=0.9\linewidth]{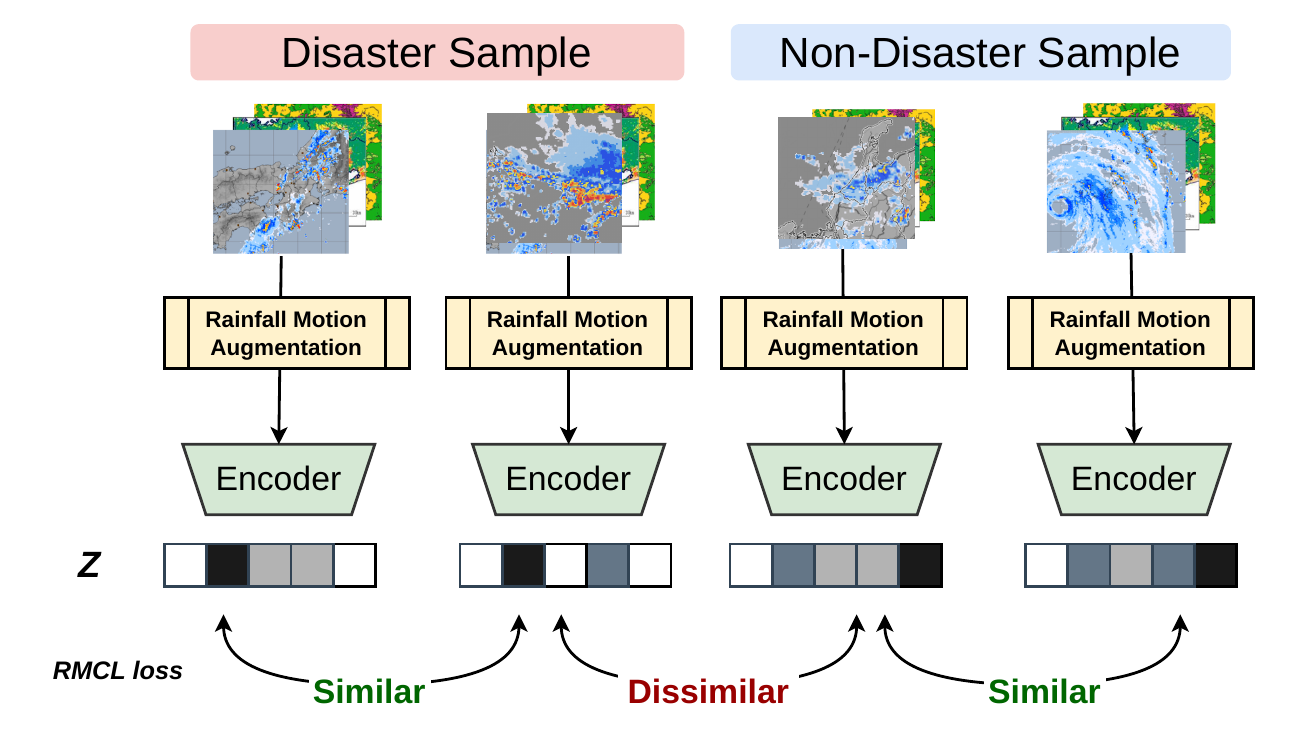}
    \caption{Rainfall Motion-Aware Contrastive Learning (RMCL) illustration.}
    \label{fig:RMCL}
\end{figure}

To learn displacement-invariant representations, we adopt a supervised contrastive learning framework~\cite{khosla2020supervised}.
The key idea is that samples corresponding to the same landslide outcome should produce similar latent representations even if the rainfall field is spatially displaced due to forecast errors as shown in figure.\ref{fig:RMCL}.
Therefore, we generate augmented rainfall samples that simulate spatial displacement and train the encoder to map both the original and perturbed samples to nearby points in the embedding space.
Given an input sample $x_i$ and its latent representation $\mathbf{z}_i = f_\theta(x_i)$, the supervised contrastive loss for an anchor $x_i$ in batch $B$ is defined as:

\begin{equation}
\mathcal{L}^{i}_{\text{RMCL}} = 
\frac{-1}{|pos(i)|}
\sum_{p \in pos(i)}
\log
\frac{
    \exp\left( \mathrm{sim}(\mathbf{z}_i, \mathbf{z}_p)/\tau \right)
}{
    \sum_{a \in B \setminus \{i\}}
    \exp\left( \mathrm{sim}(\mathbf{z}_i, \mathbf{z}_a)/\tau \right)
},
\end{equation}

where $\mathrm{sim}(\cdot,\cdot)$ denotes cosine similarity, $\tau$ is a temperature parameter, $pos(i)=\{j \mid y_j = y_i, j \neq i\}$, and $y$ denotes the landslide label.

\begin{figure}
    \centering
    \includegraphics[width=0.8\linewidth]{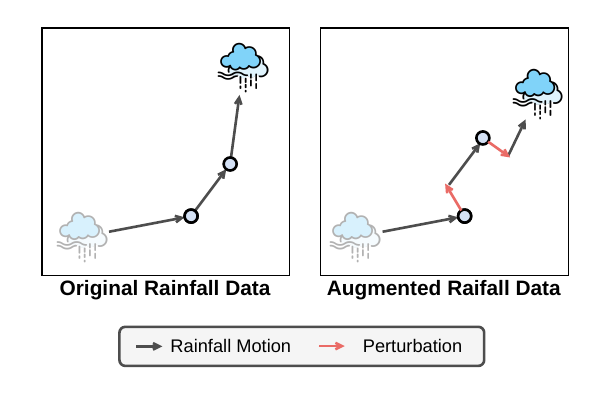}
    \caption{Proposed data augmentation design during RMCL}
    \label{fig:aug_method}
\end{figure}

The key contribution of RMCL is a data augmentation method, rainfall motion augmentation, which reflects forecast-induced displacement of rainfall fields, as illustrated in Fig.~\ref{fig:aug_method}.
Each sample consists of a rainfall field and geographically anchored terrain features, denoted as $x_i = (R_i, G_i)$.
\textcolor{black}{
To emulate the displacement of rainfall relative to terrain, augmented samples are generated by applying spatial translations only to the rainfall field $R_i$ while keeping the terrain features $G_i$ fixed.
Unlike simple augmentation strategies that translate rainfall fields independently at each time step, our method introduces temporally correlated perturbations.
Specifically, the displacement at each time step is applied relative to the displaced position from the previous step, producing sequential perturbations that mimic the accumulation of positional errors in rainfall forecasting. Consequently, the augmented data better reflects plausible forecast uncertainties and improves the robustness of landslide prediction models under operational conditions.
}
In addition, Gaussian noise sampled from a zero-mean normal distribution is added to both rainfall and terrain features.
The augmented sample is then encoded using the same feature extractor.

By optimizing the supervised contrastive objective over both original and displacement-perturbed samples sharing the same landslide label, the model learns to project displaced yet label-consistent inputs into a shared latent representation space. 
This perturbation strategy embeds robustness to forecast-induced rainfall field displacement into the learned representation, which is essential for reliable short-term landslide prediction.

\subsubsection{Landslide Predictor and Loss Function}
\label{subsec:classifier}

To predict landslides from the learned representation, we use a lightweight multi-layer perceptron (MLP) that maps the latent representation $\mathbf{z}_i = f_\theta(x_i)$ to the predicted landslide probability $\hat{y}_i \in [0,1]$. 
The predictor is trained jointly with the feature extractor in an end-to-end manner.
Since landslide events are rare, the dataset exhibits significant class imbalance. 
To mitigate this issue, we adopt focal loss~\cite{lin2017focal}, which emphasizes hard and minority samples during training. 
For binary classification, the focal loss is defined as:

\begin{equation}
\mathcal{L}_{\text{focal}} =
- \alpha (1 - \hat{y}_i)^{\gamma} y_i \log(\hat{y}_i)
- (1 - \alpha) \hat{y}_i^{\gamma} (1 - y_i) \log(1 - \hat{y}_i),
\end{equation}

where $y_i$ denotes the ground-truth label, and $\alpha$ and $\gamma$ control the class balance and focusing strength, respectively.

\section{Evaluation}

\subsection{Data Collection and Configuration}
\label{sec:data_collection_and_configuration}

To evaluate the proposed method, we collected data on landslide occurrences in Japan over two years (2021 and 2022).  
Since landslides depend on multiple factors, we focused on standard variables widely used in previous studies to construct the dataset.  
We selected 19 regions where more than three landslide events were confirmed during the observation period.
Each study region covers an area of $10 \text{ km} \times 10 \text{ km}$ and is divided into $1 \text{ km} \times 1 \text{ km}$ grid cells according to Japan’s MESH3 boundaries.

\textbf{Landslide Events:}  
We collected 253 landslide events that occurred in Japan between 2021 and 2022. Each event is annotated with the occurrence time and location (latitude and longitude up to six decimal places).  
The temporal resolution of the event is hourly, and the proposed system predicts landslide occurrences at this hourly granularity, which is sufficient for evacuation planning.

\textbf{Rainfall:}  
As shown in previous studies \cite{xie2019application}, time-series rainfall data are crucial for landslide prediction.  
The dataset includes rainfall data from January 1, 2021, to December 31, 2022.
In this study, we used radar-derived rainfall data that were corrected using ground-based gauge observations\cite{nagata2011quantitative}.

\textbf{Soil Properties:}  
Soil moisture is an important factor for landslide occurrences \cite{xie2019application, kuramoto2001study}.  
Soil slipperiness and drainage capacity significantly influence landslide risk.  
The dataset represents soil types in 10 categories, including artificial pavements.

\textbf{Vegetation:}  
Roots of plants and trees stabilize groundwater and soil, so surrounding vegetation also affects landslide risk.  
In this study, 11 vegetation categories are defined.

\textbf{Elevation and Slope Angle:}  
Elevation and slope angle are closely related to water accumulation and gravitational effects, making them important factors for landslide prediction.  
Steeper slopes are generally associated with higher disaster risk.  
Elevation is represented as a continuous value, while slope angle is classified into eight directional categories.

Each sample consists of a 48-hour rainfall sequence used for landslide prediction.
In the \textit{Forecasted-Rainfall Setting}, the sequence comprises 40 hours of observed rainfall followed by 8 hours of forecast rainfall generated using an optical-flow-based extrapolation method with the dense Lucas–Kanade algorithm \cite{bouguet2001pyramidal}, where the motion field is estimated from the three most recent rainfall fields.
In the \textit{Observed-Rainfall Setting}, the full 48-hour sequence consists solely of observed rainfall data and serves as an upper-bound reference.
A sample was labeled as positive if a landslide occurred within the subsequent eight hours; otherwise, it was labeled as negative.
To ensure the quality of negative samples, we used all data from rainy days in each region, resulting in a total of 3,567 samples in the dataset.
In this study, the AdamW optimizer was used for training, with 100 epochs for pretraining and 50 epochs for fine-tuning.

\subsection{Evaluation Metrics}
\label{sec:criteria}
Our proposed system aims to predict landslides accurately.
In disaster management systems, recall is especially important as it directly affects residents' evacuation actions.  
At the same time, maintaining precision is necessary to avoid excessive false alarms, which can undermine trust.  
Thus, we adopt \textit{precision at 80\% recall} as the main evaluation metric, and we refer to the precision at 80\% recall simply as precision hereafter.

\begin{table}[tb]
\centering
\caption{Performance comparison of the proposed and state-of-the-art methods (Precision@80\%Recall)}
\begin{threeparttable}
\begin{tabular}{lcc}
\toprule
\multirow{2}{*}{Method} & 
\multicolumn{2}{c}{Precision@80\%Recall} \\
\cmidrule(lr){2-3}
& Observed Rainfall & Forecasted Rainfall \\
\midrule
SWI-EWS \cite{kuramoto2001study} & 0.36 & 0.32 \\
\hline
CTLGNet \cite{zhao2024landslide} & 0.42 & 0.25 \\
LiteTransNet \cite{ge2024litetransnet} & 0.47 & 0.25 \\
\hline
Proposed & \textbf{0.90} & \textbf{0.69} \\
\bottomrule
\end{tabular}
\end{threeparttable}
\label{tab:results}
\end{table}

\subsection{Comparison with State-of-the-Art Methods}

To evaluate the effectiveness of the proposed system, we replaced the landslide prediction model with several state-of-the-art methods and compared their prediction performance under the same rainfall forecasting conditions.
Specifically, we integrated SWI-EWS \cite{kuramoto2001study}, CTLGNet \cite{zhao2024landslide}, and LiteTransNet \cite{ge2024litetransnet} into the proposed LEWS as alternative landslide prediction models.
The rainfall forecasts generated by the rainfall forecasting model were provided as inputs to each method, and the resulting landslide prediction performance was evaluated.

\textbf{SWI-EWS} \cite{kuramoto2001study}: This method employs a fixed-parameter three-layer tank model to compute the Soil Water Index (SWI) from rainfall.
It then statistically evaluates rainfall abnormality based on historical records to predict landslide occurrence.
This approach is currently operated in practice in Japan.

\textbf{CTLGNet} \cite{zhao2024landslide}: A deep learning model that combines CNN and Transformer architectures to capture spatio-temporal patterns of landslide occurrence.

\textbf{LiteTransNet} \cite{ge2024litetransnet}: A lightweight Transformer-based model that uses localized self-attention to capture the varying importance of historical timestamps in landslide displacement prediction.

As described in the Evaluation Metrics section, model performance was assessed using precision at 80\% recall.
Table~\ref{tab:results} reports these metrics for all methods.
The results demonstrate that the proposed method outperforms all baseline methods in the observed-rainfall setting.
Furthermore, in forecast rainfall settings, the proposed system also outperforms other state-of-the-art landslide prediction methods, achieving a 37\% improvement in precision compared with the strongest alternative.
This result demonstrates the effectiveness of RMCL-based representation learning for robust landslide prediction under rainfall forecast uncertainty.

\subsection{Effect of Rainfall-Motion-Aware Contrastive Learning}
To evaluate the effectiveness of Rainfall-Motion-Aware Contrastive Learning (RMCL), we conducted an ablation study comparing three training strategies:
(i) standard supervised learning using focal loss (End-to-end),
(ii) supervised training using forecast rainfall as input (End-to-end w/ Forecast Rain), and
(iii) the proposed RMCL-based training.

The comparison was performed under two scenarios: using observed rainfall and using forecast rainfall.
The results show that the model trained with RMCL consistently achieved the highest prediction accuracy in both scenarios.
In particular, the proposed method improved accuracy by approximately 50\% compared with the two baseline training strategies under the forecast rainfall setting.
These results indicate that RMCL effectively improves the robustness of the landslide prediction model, especially under operational conditions where forecast rainfall is used as input.
Overall, the findings demonstrate that RMCL-based training is effective for achieving accurate future landslide prediction in practical early warning settings that rely on rainfall forecasts.

\begin{figure}[tb]
    \centering
    \includegraphics[width=0.85\linewidth]{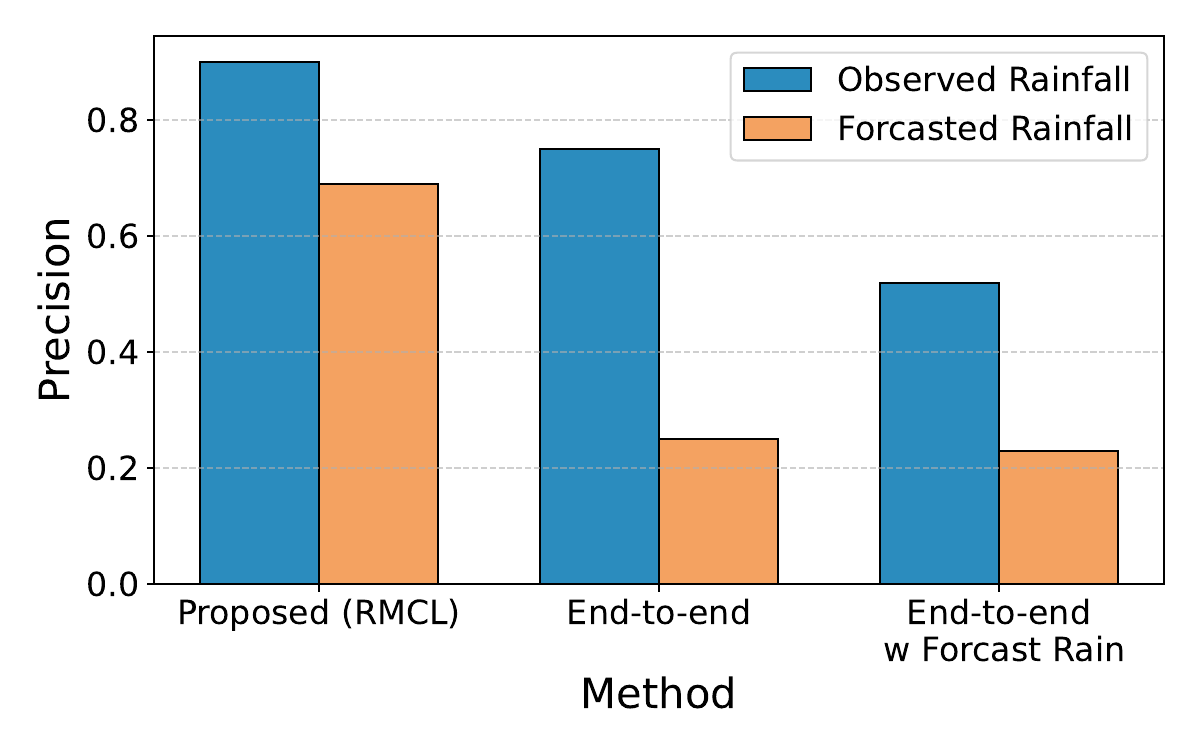}
    \vspace{-0.5cm}
    \caption{Ablation study on RMCL}
    \label{fig:ablation_study_leave-k-out}
    \vspace{-0.5cm}
\end{figure}

\section{Conclusion}

In this study, we addressed landslide prediction under rainfall forecast uncertainty, where spatial displacement of rainfall fields can affect prediction accuracy. 
We proposed a novel LEWS that is robust to the spatial displacement of rainfall fields caused by rainfall forecasting.
To improve robustness to rainfall field displacement, the proposed system introduces a temporally correlated rainfall field perturbation process and learns stable representations from rainfall and terrain data using RMCL.
This design improves the robustness of learned representations to displacement in rainfall field motion and enhances the reliability of landslide prediction under operational conditions.
Experiments on data from 19 regions across Japan show up to a 37\% improvement in precision over conventional deep learning and statistical baselines, demonstrating the effectiveness of modeling rainfall field displacement and data imbalance in short-term landslide prediction. 
These results indicate that modeling rainfall field motion and learning stable representations through RMCL are important for capturing early indicators of landslide occurrence.
Future work will extend the framework to generalize the approach to other rainfall-induced hazards, including floods and inundation events.

\bibliographystyle{IEEEtran}
\bibliography{sample-base}

\end{document}